\documentclass[letterpaper, 10 pt, conference]{ieeeconf}
\IEEEoverridecommandlockouts 
\let\labelindent\relax
\usepackage{enumitem}
\usepackage{multicol}

\usepackage{bm}
\usepackage{amsmath}
\usepackage{amssymb}
\usepackage{mathtools} 
\usepackage{amsfonts}
\usepackage{comment}
\usepackage{graphicx}
\usepackage[dvipsnames]{xcolor}
\usepackage{tikz}

\usepackage{gensymb}
\usepackage[linesnumbered,ruled,vlined]{algorithm2e}

\usepackage{colortbl}

\usepackage{todonotes}
\usepackage{siunitx}
\usepackage{subcaption}
\usepackage{booktabs}
\usepackage{tabularx}
\usepackage{url}

\newtheorem{assumption}{Assumption}

\title{\LARGE \bf
Old Experience, New Lessons: Relabeling Autonomous Robot Rollouts in Interactive Imitation Learning}

\newif\ifanonymous
\anonymousfalse
\ifanonymous
  \author{Anonymous Authors}
\else
  \author{Zhaoting Li$^{1}$, Cosimo Della Santina$^{1}$, and Jens Kober$^{2}$%
  \thanks{$^{1}$Z. Li and C. Della Santina are with Delft University of Technology, The Netherlands.
  $^{2}$J. Kober is with University of Stuttgart, Germany.}%
  }
\fi

\begin{document}
\bstctlcite{IEEEexample:BSTcontrol}

\maketitle
\ifanonymous\else
  \setcounter{footnote}{2} 
\fi
\thispagestyle{empty}
\pagestyle{empty}

\begin{abstract}
Interactive imitation learning enables robots to learn from human
interventions during deployment. These interactions naturally produce sparse
human corrections and abundant autonomous robot experience without human
action labels. Yet existing methods either leave this experience unused for
action supervision or imitate historical robot actions, which can reinforce
suboptimal behavior. To address this limitation, we propose
Set-valued Relabeling of Autonomous experience (SeRA), a framework that learns from
autonomous robot experience through set-valued supervision of policy
relabels. Later policy checkpoints relabel this experience, and agreement
across checkpoints filters the relabels. Since agreement does not guarantee
correctness, SeRA constructs desired action sets around accepted relabels.
Following Set-Supervised Diffusion Policy (SDP), it trains an action-chunking
diffusion policy to increase the probability of actions within these sets.
A step-masked denoising loss combines accepted relabels with
available human corrections in partially supervised chunks. Experiments
across five simulated manipulation tasks in online and offline settings show
that SeRA improves policy performance over SDP and outperforms baselines that imitate
historical robot actions or exact relabels. On a real-robot tomato-picking
task, SeRA achieves 80\% success, compared with 40\% for SDP, using the
same human feedback and training budget. These results demonstrate that
relabeled autonomous robot experience can complement human corrections to
improve policy performance.
\end{abstract}

\section{Introduction}
\label{sec:intro}

Learned manipulation policies remain vulnerable to failures around contacts,
task bottlenecks, and states outside the demonstration distribution.
Interactive imitation learning (IIL) addresses these failures by allowing a human
teacher to intervene during deployment and provide
corrections~\cite{2019_HG_DAgger,hoque2022thriftydagger}. These interactions
naturally produce two streams of data: sparse human corrections and abundant
autonomous robot experience without human action labels. Learning only from
the corrections leaves this experience unused for action supervision. We
instead treat it as unlabeled data and seek to learn from it alongside human
corrections.

Prior work has sought to exploit autonomous robot experience by retaining
historical robot actions as imitation targets. IWR imitates these actions
alongside human corrections~\cite{mandlekar2020iwr}; CEILing uses additional evaluative
feedback to select useful behavior~\cite{2022_RAL_Correct_me_If_Wrong}; and
Sirius downweights pre-intervention actions using intervention
context~\cite{liu2023sirius}. However, autonomous actions need not be optimal,
even when they do not trigger an intervention or occur in a successful
trajectory. Retaining them as targets can therefore reinforce suboptimal
behavior. This raises a central question: how can autonomous robot experience
support policy learning without imitating imperfect historical robot actions?

\begin{figure}[t]
  \centering
  \includegraphics[width=0.98\columnwidth]{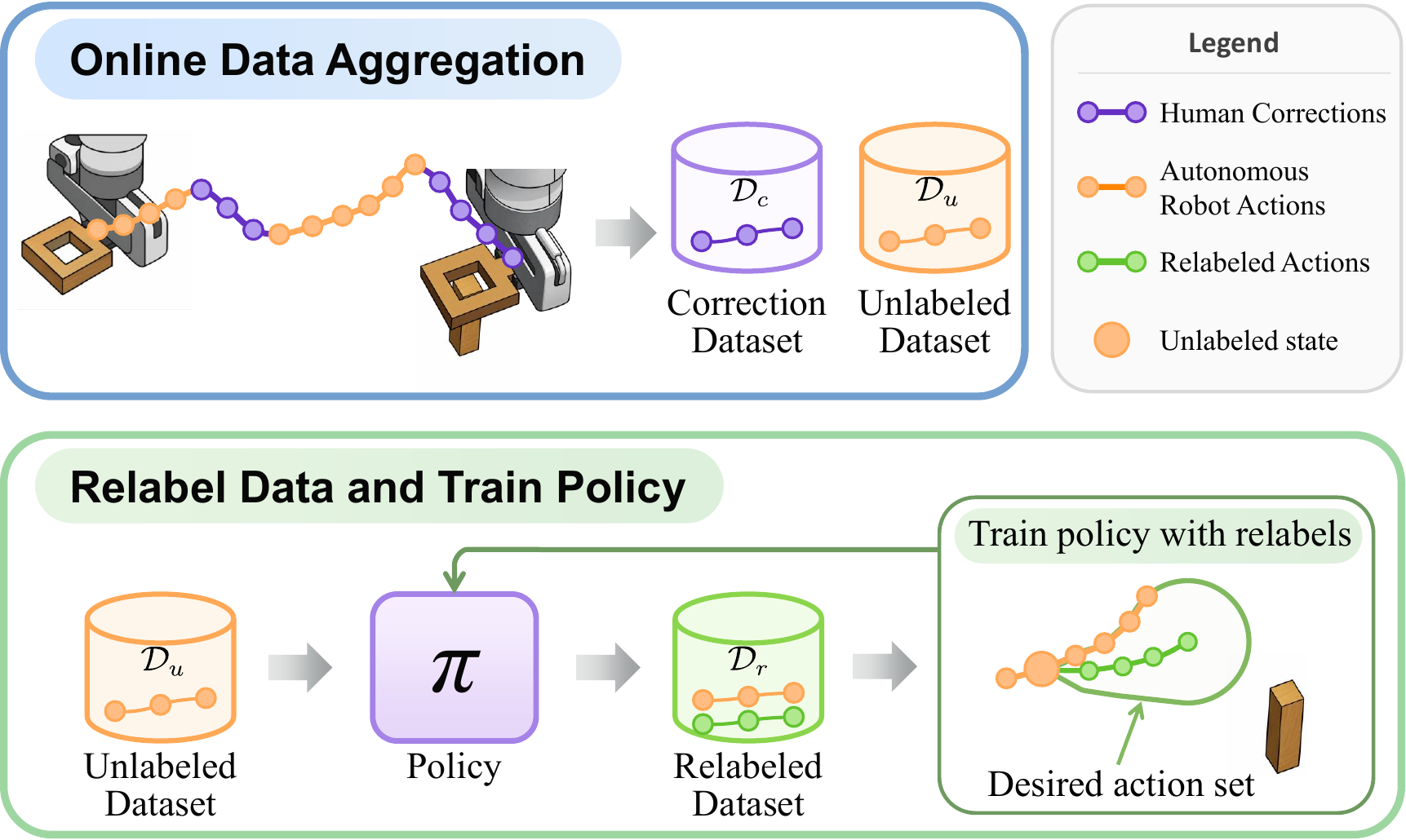}
  \caption{SeRA framework. Top: during online interaction, human corrections are
  stored in a correction dataset and autonomous robot experience in an unlabeled
  dataset. Bottom: a later, improved policy $\pi$ relabels the unlabeled
  observations with new action chunks (green) in place of the historical robot
  actions (orange). Relabels accepted by checkpoint agreement yield desired
  action sets that supervise training alongside the corrections.}
  \label{fig:cover}
\end{figure}

Our key insight is that later policies incorporate additional human
corrections and may generalize these corrections to observations collected
earlier. We use the latest policy to relabel autonomous robot experience
and accept these relabels based on agreement across later policy checkpoints.
Yet agreement alone does not
guarantee correctness: multiple checkpoints can share the same error.
Directly imitating even these accepted relabels can reinforce prediction
errors through confirmation bias~\cite{arazo2020pseudolabeling}. Learning
from this experience therefore requires supervision that tolerates imperfect
relabels.

In this work, we introduce Set-valued Relabeling of Autonomous experience
(SeRA),\footnote{Webpage: \url{https://sera-webpage.github.io/}} a
framework that combines policy relabeling with set-valued supervision for
action-chunking diffusion policies. Building on Set-Supervised Diffusion
Policy (SDP)~\cite{Li2026SDP}, we supervise each accepted relabel with a
desired action set: a region of action space centered at the relabel, whose
radius is scaled by its distance from the historical robot action. Rather
than requiring the relabel itself to be optimal, we assume that the desired
set contains an optimal action. Following SDP, we train the policy
to increase the probability of actions within the desired sets. A further
challenge arises because intervention boundaries and relabel filtering leave
action chunks only partially supervised. Discarding these chunks would
waste usable feedback. We therefore introduce a step-masked denoising loss that
combines retained human corrections and accepted relabels within a chunk,
learning from supported steps while excluding unsupported steps.

We evaluate SeRA in online and offline learning across five simulated
manipulation tasks and in a real-robot tomato-picking task. Comparisons with
methods that imitate historical robot actions and with exact relabel
imitation examine how autonomous robot experience contributes to policy learning,
while ablations assess each component of SeRA. Across these
experiments, SeRA consistently outperforms the baselines.

In summary, this work makes the following contributions:
\begin{itemize}[leftmargin=*]
    \item Set-valued supervision of policy relabels to learn from autonomous
    robot experience, using checkpoint agreement to accept relabels and
    desired action sets to relax exact matching of these imperfect relabels.
    \item A step-masked denoising loss that learns from partially supervised action
    chunks by combining retained human corrections and accepted relabels.
    \item An evaluation in online and offline simulation and on a real robot,
    with baseline comparisons and component ablations.
\end{itemize}

\section{Related Work}
\label{sec:related_work}

\subsection{Interactive Imitation and Autonomous Robot Experience}

DAgger addresses covariate shift through expert labels at learner-visited
states~\cite{2011_DAgger}. HG-DAgger and ThriftyDAgger refine intervention
gating~\cite{2019_HG_DAgger,hoque2022thriftydagger}, while CLIC and SDP
learn desired action sets from teacher corrections paired with robot
actions~\cite{2025_CLIC,Li2026SDP}. Autonomous robot experience can support
policy learning: IWR balances imitation of robot actions and
human corrections~\cite{mandlekar2020iwr}, Sirius prioritizes interventions
and downweights pre-intervention actions~\cite{liu2023sirius}, and CEILing
combines corrective and evaluative feedback~\cite{2022_RAL_Correct_me_If_Wrong}.
These approaches retain selected historical robot actions as supervision;
we instead generate new action targets at the recorded observations.

Other methods exploit autonomous robot experience with additional reward
information. Offline reinforcement learning (RL) improves policies from fixed
datasets with recorded
rewards~\cite{levine2020offline,wang2023diffusionql,zheng2023ssorl}, and rewards
can also be learned from limited
supervision~\cite{zolna2020oril,konyushkova2020semisupervisedreward,park2022surf}.
Beyond rewards, implicit quality signals can be inferred from intervention
decisions~\cite{2022_Expert_Intervention_Learning,korkmaz2025mile}, and actions
can be relabeled using verbal corrections~\cite{liu2023olaf}. SeRA instead
requires no reward signal and relabels autonomous robot experience using later
policies trained with human corrections.

Our work is most closely related to SDP~\cite{Li2026SDP}, which constructs desired
action sets from human corrections paired with robot actions. SeRA
extends this supervision to policy-generated
labels and combines accepted labels with partial human corrections within
action chunks through a step-masked denoising loss.

\subsection{Semi-Supervised Learning}

Offline imitation methods identify expert trajectories through
positive-unlabeled learning~\cite{wang2023pubc} or exploit mixed-quality
demonstrations through weighting, selection, self-supervision, and
representation learning~\cite{xu2022dwbc,wu2025imperfect,yue2024diverse,yang2022trail}.
We treat autonomous robot experience as unlabeled with respect to desired actions.
Pseudo-labeling supplies targets but can reinforce errors through confirmation
bias~\cite{arazo2020pseudolabeling}. Uncertainty filtering and consistency
regularization improve their
reliability~\cite{rizve2021defense,sohn2020fixmatch,berthelot2020remixmatch},
while Temporal Ensembling and Mean Teacher stabilize targets over
training~\cite{laine2017temporal,tarvainen2017mean}.

Continuous-output methods account for pseudo-label uncertainty through
ensembling and weighting~\cite{dai2023ucvme,sun2025heteroscedastic};
set-valued targets have also been studied over class
distributions~\cite{lienen2021credal}. SeRA filters latest-policy
relabels by agreement across later interaction-round checkpoints. Since
agreement does not guarantee correctness, it relaxes exact relabel matching
through desired sets for continuous action chunks. A step-masked denoising loss combines
relabeled and partial human supervision, excluding unsupported steps.

\section{Preliminary}
\label{sec:preliminary}

\subsection{Interactive imitation learning}
\label{sec:prelim_iil}
In the IIL framework, a robot executes its policy
while a human teacher can intervene, and the policy is trained on the collected
data. Let $\mathbf{O}_t$ be the policy
observation at time $t$, $\mathbf{a}_t^r\in\mathcal{A}$
the robot action from policy $\pi$, and
$\mathbf{a}_t^h\in\mathcal{A}\cup\{\varnothing\}$ the human corrective action
from expert policy $\pi^*$. We denote trajectories as sequences
$\tau=(\mathbf{O}_1,\mathbf{a}_1^r,\mathbf{a}_1^h,\mathbf{O}_2,\mathbf{a}_2^r,\mathbf{a}_2^h,\ldots)$
together with an intervention indicator $m_t\in\{0,1\}$, where
$\mathbf{a}_t^h=\varnothing$ when $m_t=0$.

The policy predicts action chunks
$\mathbf{A}_t=[\mathbf{a}_t,\ldots,\mathbf{a}_{t+T-1}]$ of length $T$, where
$\mathbf{a}_{t+j}$ is the action at step $j$ of the chunk. We write
$\mathbf{A}_t^r$ and $\mathbf{A}_t^h$ for the robot and human action chunks
starting at $t$. The human action chunk $\mathbf{A}_t^h$ is partially labeled
if a human action is missing at some
step. It is unlabeled if a human action is missing at every step. If a
human action is available at every step, the chunk is complete
teacher-labeled. Such chunks form the correction dataset
\begin{equation}
    \mathcal{D}_c
    \coloneqq
    \Bigl\{
        (\mathbf{O}_t,\mathbf{A}_t^r,\mathbf{A}_t^h)
        \Bigm|
        \prod\nolimits_{j=0}^{T-1}m_{t+j}=1
    \Bigr\}.
    \label{eq:intervention_dataset}
\end{equation}
Because the teacher intervenes at every step of these chunks,
$\mathbf{A}_t^r$ in $\mathcal{D}_c$ is queried from $\pi$ but not executed.

\subsection{Set-valued supervision}

CLIC introduces set-valued action supervision for corrective feedback,
constructing desired action sets from a positive action $\mathbf{a}^{+}$ and a
negative action $\mathbf{a}^{-}$ rather than imitating a single action
label~\cite{2025_CLIC}.
SDP extends this formulation to action-chunking
diffusion policies~\cite{Li2026SDP}. Given a distance
$D(\mathbf{a}_1,\mathbf{a}_2)=\lVert\mathbf{a}_1-\mathbf{a}_2\rVert_2$, the
corresponding single-step desired action set is
\begin{equation}
    \begin{aligned}
        &\widehat{\mathcal{A}}^{\mathrm{ss}}_{t}
\vcentcolon=
\widehat{\mathcal{A}}^{\mathrm{ss}}(\mathbf a_{t}^-, \mathbf a_{t}^+) \\
        &\quad=
        \left\{
            \mathbf{a}\in\mathcal{A}
            \,\middle|\,
            D(\mathbf{a},\mathbf{a}_{t}^{+})
            \le rD(\mathbf{a}_{t}^{+},\mathbf{a}_{t}^{-})
        \right\},
        \ r\in[0,1].
    \end{aligned}
    \label{eq:sdp_single_step_set}
\end{equation}
Thus, the desired set is a ball centered at the positive action; the
positive--negative distance sets its scale, and $r$ is the radius ratio.
Behavioral cloning is recovered at $r=0$, when the set reduces to
$\{\mathbf{a}^{+}\}$.

For a contrastive pair of action chunks
$\mathbf{A}_t^{-}=[\mathbf{a}_t^{-},\ldots,\mathbf{a}_{t+T-1}^{-}]$ and
$\mathbf{A}_t^{+}=[\mathbf{a}_t^{+},\ldots,\mathbf{a}_{t+T-1}^{+}]$, SDP
constructs one such ball at every step. The chunk-level desired set is
the sequence of single-step desired action sets,
\begin{equation}
    \begin{aligned}
        \widehat{\mathcal{A}}_t
        (\mathbf{A}_t^{-},\mathbf{A}_t^{+}) \vcentcolon=[\widehat{\mathcal{A}}^{\mathrm{ss}}_t, \widehat{\mathcal{A}}^{\mathrm{ss}}_{t+1}, \dots, \widehat{\mathcal{A}}^{\mathrm{ss}}_{t+T-1}].
    \end{aligned}
    \label{eq:sdp_chunk_set}
\end{equation}
Equivalently, a candidate chunk
$\mathbf{A}_t=[\mathbf{a}_t,\ldots,\mathbf{a}_{t+T-1}]$ belongs to
$\widehat{\mathcal{A}}_t$ if every component $\mathbf{a}_{t+j}$ belongs to its
corresponding single-step set in Eq.~\eqref{eq:sdp_single_step_set}.

SDP obtains a clean desired chunk $\mathbf{A}_t^0$ from the constrained target distribution
$\pi^{\mathrm{target}}(\mathbf{A}\mid\mathbf{O}_t)
\propto\mathbb{I}[\mathbf{A}\in\widehat{\mathcal{A}}_t]
\pi_\theta(\mathbf{A}\mid\mathbf{O}_t)$, and trains $\pi_\theta$ with the denoising loss
\begin{equation}
    \ell_t^{\mathrm{SDP}}(\theta)
    \coloneqq
    \mathbb{E}_{q,\boldsymbol{\epsilon},\mathbf{A}_t^0}
    \bigl[
        \lVert
            \boldsymbol{\epsilon}
            -\epsilon_\theta(\mathbf{O}_t,\mathbf{A}_t^q,q)
        \rVert_F^2
    \bigr].
    \label{eq:sdp_loss}
\end{equation}
Here, $\mathbf{A}_t^q$ is the noised version of $\mathbf{A}_t^0$ at diffusion
step $q$, and $\boldsymbol{\epsilon}$ is the sampled diffusion noise.

\section{Method}
\label{sec:method}

\begin{figure*}[t] \centering \includegraphics[width=0.99\textwidth]{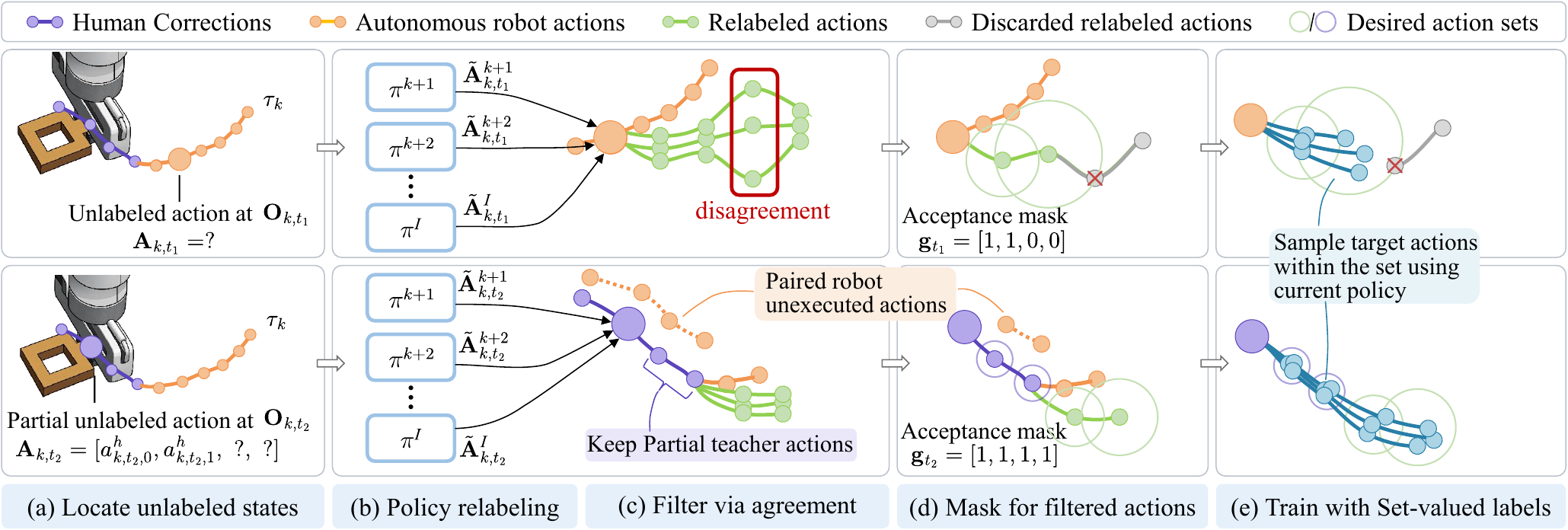} \caption{Overview of SeRA. Later policy checkpoints relabel missing teacher actions, with checkpoint agreement determining the acceptance mask. Set-valued supervision uses relabels up to the first disagreement for unlabeled chunks (top) and combines accepted relabels with initial teacher actions for partially labeled chunks (bottom).} \label{fig:chunk_alignment} \end{figure*}

IIL yields sparse human corrections and abundant
autonomous robot experience without human action labels. We treat this
experience as unlabeled data and use later policies to relabel earlier
observations. We call the resulting method
Set-valued Relabeling of Autonomous experience (SeRA). It has three components:
cross-round policy relabeling filters the relabels by checkpoint agreement
(Section~\ref{sec:relabel}); set-valued targets relax exact matching of
those relabels 
(Section~\ref{sec:set_target}); and a step-masked denoising loss combines human and relabeled
supervision within action chunks (Section~\ref{sec:masked}).
Fig.~\ref{fig:chunk_alignment} and Algorithm~\ref{alg:method} illustrate the procedure.

\subsection{Problem setup}
\label{sec:setup}

We consider learning a policy from both human corrections and autonomous
robot experience collected in IIL (Section~\ref{sec:prelim_iil}).
Learning proceeds in rounds, each ending with policy updates that
produce a new checkpoint; let $i$ index the rounds and $I$ denote the current
round. In online learning, each round $i$ starts with an interaction phase,
in which policy $\pi^i$ collects a trajectory, before the updates produce
$\pi^{i+1}$. After the interaction phase of round $I$, frozen checkpoints
$\{\pi^i\}_{i\le I}$ and trajectories $\{\tau_k\}_{k\le I}$ are available,
where $\tau_k$ is collected by $\pi^k$. In offline learning, all trajectories
$\{\tau_k\}_{k=1}^{K}$ are collected before training, so each round consists
only of policy updates.
We use $k$ to index trajectories and $i$ to index checkpoints.
Let $\mathbf{O}_{k,t}$, $m_{k,t}$, $\mathbf{A}_{k,t}^r$, and
$\mathbf{A}_{k,t}^h$ denote the observation, intervention indicator, and robot
and human action chunks at time $t$ of $\tau_k$
(Section~\ref{sec:prelim_iil}). Trajectory $\tau_k$ has length $H_k$, so valid
length-$T$ chunks start at $\mathcal{T}_k\coloneqq\{1,\ldots,H_k-T+1\}$.
Complete teacher-labeled chunks form $\mathcal{D}_c$
(Eq.~\eqref{eq:intervention_dataset}); the remaining chunks contain at least
one missing human action. We denote the observations at which these chunks
start as unlabeled states (Fig.~\ref{fig:chunk_alignment}(a)). In $\tau_k$,
their time indices form
\begin{equation}
    \mathcal{N}_k
    \coloneqq
    \Bigl\{
        t\in\mathcal{T}_k
        \Bigm|
        \prod\nolimits_{j=0}^{T-1}m_{k,t+j}=0
    \Bigr\}.
    \label{eq:nonintervention_starts}
\end{equation}
Our goal is to train $\pi^{I+1}$ on both $\mathcal{D}_c$ and the unlabeled
dataset
$\mathcal{D}_u\coloneqq\{(\mathbf{O}_{k,t},\mathbf{A}_{k,t}^r,\mathbf{A}_{k,t}^h)
\mid t\in\mathcal{N}_k\}$. The checkpoints
$\{\pi^i\}_{i\le I}$ construct supervision at the unlabeled states of earlier
trajectories by relabeling their missing human actions.

\subsection{Cross-round policy relabeling}
\label{sec:relabel}

Policies trained after $\tau_k$ was collected incorporate additional human
corrections and may provide improved relabels at its unlabeled states
$\mathbf{O}_{k,t}$, $t\in\mathcal{N}_k$. For each trajectory $\tau_k$
collected before round $I$, we query the later checkpoints at each unlabeled
state $\mathbf{O}_{k,t}$ (Fig.~\ref{fig:chunk_alignment}(b)):
\begin{equation}
    \tilde{\mathbf{A}}_{k,t}^{i}
    = [\tilde{\mathbf{a}}_{k,t}^{i},\ldots,
       \tilde{\mathbf{a}}_{k,t+T-1}^{i}]
    \sim\pi^i(\cdot\mid\mathbf{O}_{k,t}),
    \label{eq:later_policy_query}
\end{equation}
where $\tilde{\mathbf{a}}_{k,t+j}^{i}$ is the action predicted for step $j$ of
the chunk, and $i\in\mathcal{I}_k$ indexes the later checkpoints, trained
after $\tau_k$ was collected, up to the most recent one $\pi^I$. In online learning, $\mathcal{I}_k=\{k+1,\ldots,I\}$; for
example, at round $I=20$, $\pi^6,\ldots,\pi^{20}$ are queried at the
unlabeled states of $\tau_5$. In offline learning,
$\mathcal{I}_k=\{1,\ldots,I\}$ for every trajectory; for example, at round
$I=20$, $\pi^1,\ldots,\pi^{20}$ are queried at the unlabeled states of every
trajectory. The latest prediction $\tilde{\mathbf{A}}_{k,t}^{I}$
supplies the relabel; agreement across checkpoints determines its acceptance.
This use of temporal consistency is motivated by Temporal Ensembling and Mean
Teacher, which use a model's training history to provide stable supervision
for unlabeled data~\cite{laine2017temporal,tarvainen2017mean}.

\paragraph{Consistency filtering}
Since individual checkpoints may produce erroneous relabels, we use low
cross-checkpoint variance as a proxy for relabel reliability.
For each step $j$, we compute the total variance of the predictions across the
$n_k\coloneqq|\mathcal{I}_k|$ later checkpoints:
\begin{equation}
    v_{k,t+j}
    \coloneqq
    \frac{1}{n_k-1}\sum\nolimits_{i\in\mathcal{I}_k}
    \left\lVert
        \tilde{\mathbf{a}}_{k,t+j}^{i}-\bar{\mathbf{a}}_{k,t+j}
    \right\rVert_2^2,
    \label{eq:step_consistency}
\end{equation}
where $\bar{\mathbf{a}}_{k,t+j}$ is the mean of the predicted actions
$\tilde{\mathbf{a}}_{k,t+j}^{i}$ over the later checkpoints.
The relabel at step $j$ passes the consistency test if enough later
checkpoints are available and their predictions agree:
\begin{equation}
    c_{k,t+j}\coloneqq
    \begin{cases}
        \mathbb{I}[v_{k,t+j}<\delta], & n_k\ge M,\\
        0, & n_k<M.
    \end{cases}
    \label{eq:consistency_test}
\end{equation}
where $M\ge2$ is the minimum number of later checkpoints and $\delta$ is the
variance threshold.
In an unlabeled chunk, the relabel at step $j$ is accepted only if the relabels
at steps $0,\ldots,j$ all pass the consistency test. A failed test thus rejects
its step and all later steps of the chunk. For example, in the top row of
Fig.~\ref{fig:chunk_alignment}(c), disagreement at step $2$ also rejects
step $3$, although step $3$ passes individually.
In practice, earlier checkpoints have already been queried in previous rounds,
so round $I$ only queries the current checkpoint $\pi^I$ and updates the
variance incrementally.

Hereafter, we omit the trajectory index $k$ and latest-checkpoint superscript $I$
when the context is clear, and write $\tilde{\mathbf{a}}_{t+j}$ for the relabel
at step $j$ of the chunk starting at $t$.

\subsection{Relabels as set-valued targets}
\label{sec:set_target}

Cross-round agreement selects stable relabels, but shared errors can persist
across checkpoints. Directly imitating accepted relabels can therefore
reinforce their errors through confirmation bias~\cite{arazo2020pseudolabeling}.
We instead use each accepted relabel as a positive action paired
with the historical robot action, following the set-valued supervision of
  SDP~\cite{Li2026SDP}.

Let $\widehat{\mathcal{A}}^{\mathrm{ss}}(\mathbf{a}^-,\mathbf{a}^+;r)$ denote
the single-step desired set in Eq.~\eqref{eq:sdp_single_step_set}. An accepted
relabel defines the set
$\widehat{\mathcal{A}}^{\mathrm{ss}}
(\mathbf{a}_{t+j}^r,\tilde{\mathbf{a}}_{t+j};r^N)$
(Fig.~\ref{fig:chunk_alignment}(e)),
centered at the relabel with radius
$r^N\lVert\tilde{\mathbf{a}}_{t+j}-\mathbf{a}_{t+j}^r\rVert_2$.
We set $r^N=1$ to account for relabel uncertainty, so the radius equals the
distance from the relabel to the historical robot action. Rather than assume
that the relabel is optimal, we make this assumption.

\begin{assumption}
\label{asm:relabel_set}
At every accepted relabeled step $j$ of a chunk starting at $t$, the desired
set $\widehat{\mathcal{A}}^{\mathrm{ss}}
(\mathbf{a}_{t+j}^r,\tilde{\mathbf{a}}_{t+j};r^N)$ contains the action at step
$j$ of an optimal chunk at $\mathbf{O}_t$.
\end{assumption}

This assumption is weaker than requiring the relabel itself to be optimal
($r^N=0$). Set-valued supervision thus allows the policy to learn from
imperfect relabels without requiring it to reproduce them exactly.

\subsection{Learning from partially supervised chunks}
\label{sec:masked}

Complete teacher-labeled chunks in $\mathcal{D}_c$
(Eq.~\eqref{eq:intervention_dataset}) use the SDP loss in Eq.~\eqref{eq:sdp_loss}, with
robot and human actions as negative and positive actions and radius ratio
$r^h$. A chunk that is not complete teacher-labeled can still contain human
actions, for example when an intervention ends within the chunk
(Fig.~\ref{fig:chunk_alignment}(c), bottom row). Rather than discarding these partial human
actions, we combine them with accepted relabels in a single chunk.

\paragraph{Selecting supervision}
For a chunk starting at $t\in\mathcal{N}_k$, let $h_t$ be the number of
consecutive human actions from its start, up to the first missing human
action.
These first $h_t$ steps retain their human actions.
The remaining steps use relabels, accepted from step $h_t$ up to the first
failed consistency test. The resulting acceptance mask is
$\mathbf{g}_t=[g_t,\ldots,g_{t+T-1}]$
(Fig.~\ref{fig:chunk_alignment}(d)), where
\begin{equation}
    g_{t+j}\coloneqq
    \begin{cases}
        1, & j<h_t,\\
        \prod\nolimits_{\ell=h_t}^{j}c_{t+\ell}, & j\ge h_t.
    \end{cases}
    \label{eq:relabel_acceptance_vector}
\end{equation}
An unlabeled chunk has $h_t=0$, so its mask accepts relabels up to
the first failed test, as in Section~\ref{sec:relabel}. Chunks whose mask
accepts the first step form the relabeled dataset
\begin{equation}
    \begin{aligned}
        \mathcal{D}_r
        \coloneqq
        \bigl\{
        &(\mathbf{O}_{k,t},\mathbf{A}_{k,t}^r,\mathbf{A}_{k,t}^h,
        \tilde{\mathbf{A}}_{k,t}^{I},\mathbf{g}_{k,t})\\
        &\bigm|
        \mathcal{I}_k\neq\emptyset,\ t\in\mathcal{N}_k,\ g_{k,t}=1
        \bigr\},
    \end{aligned}
    \label{eq:relabeled_dataset}
\end{equation}
which retains the robot action chunks for the negative actions, and the
partial human action chunks and relabels for the positive actions.
At each accepted step, the positive action and radius ratio are
\begin{equation}
    (\mathbf{a}_{t+j}^+,r_{t+j})
    \coloneqq
    \begin{cases}
        (\mathbf{a}_{t+j}^h,\,r^h), & j<h_t,\\
        (\tilde{\mathbf{a}}_{t+j},\,r^N), & j\ge h_t.
    \end{cases}
    \label{eq:positive_action}
\end{equation}
Human corrections use narrower sets, $r^h<r^N$, since they are more reliable
than relabels.

\paragraph{Step-masked denoising loss}
Steps from the first failed consistency test onward are masked
($g_{t+j}=0$) and provide no supervision. The chunk-level desired set
$\widehat{\mathcal{A}}_t$ in Eq.~\eqref{eq:sdp_chunk_set} therefore constrains
each accepted step ($g_{t+j}=1$) to $\widehat{\mathcal{A}}^{\mathrm{ss}}
(\mathbf{a}_{t+j}^r,\mathbf{a}_{t+j}^+;r_{t+j})$ and leaves masked steps
unconstrained. As in SDP, we draw a clean target chunk $\mathbf{A}_t^0$ from
the target distribution $\pi^{\mathrm{target}}$ over this set, which preserves
the current policy's relative probabilities within the set. However, the
sampled target chunk also contains actions at unconstrained steps, where no trusted
supervision is available. An unmasked denoising loss would nevertheless train
the policy to reproduce these sampled actions. We therefore apply the diagonal mask
$\mathbf{C}_t\coloneqq\operatorname{diag}(\mathbf{g}_t)$ to the denoising
residual, restricting supervision to accepted steps:
\begin{equation}
    \ell_t^{\mathrm{M}}(\theta)
    \coloneqq
    \mathbb{E}_{q,\boldsymbol{\epsilon},\mathbf{A}_t^0}
    \bigl[
        \lVert
            \mathbf{C}_t(
            \boldsymbol{\epsilon}
            -\epsilon_\theta(\mathbf{O}_t,\mathbf{A}_t^q,q)
            )
        \rVert_F^2
    \bigr],
    \label{eq:masked_sdp_loss}
\end{equation}
where $q$, $\mathbf{A}_t^q$, and $\boldsymbol{\epsilon}$ are defined as in
Eq.~\eqref{eq:sdp_loss}. Without masked steps ($\mathbf{g}_t=\mathbf{1}$),
Eq.~\eqref{eq:masked_sdp_loss} reduces to the SDP loss.

Following SDP~\cite{Li2026SDP}, learning from the desired sets of multiple
chunks can progressively narrow the acceptable behavior: policy updates from
multiple samples jointly enforce compatible constraints, and policy
generalization extends this effect to similar observations.

\paragraph{Full objective}
We train the policy on both datasets:
\begin{equation}
    \mathcal{L}(\theta)
    \coloneqq
    \mathbb{E}_{\mathcal{D}_c}
    \bigl[\ell_t^{\mathrm{SDP}}(\theta)\bigr]
    +\beta\,
    \mathbb{E}_{\mathcal{D}_r}
    \bigl[\ell_t^{\mathrm{M}}(\theta)\bigr],
    \label{eq:full_objective}
\end{equation}
where each expectation is over chunks in its dataset, and
$\beta>0$ balances corrective and relabeled supervision.

\begin{algorithm}[t]
\caption{SeRA}
\label{alg:method}
\DontPrintSemicolon
\KwIn{Initial policy $\pi^0$, pre-collected trajectories}
Construct $\mathcal{D}_c$ by Eq.~\eqref{eq:intervention_dataset} from the
trajectories\;
\For{round $I=0,1,\ldots$}{
    \If{online learning}{
        Deploy $\pi^I$, receive corrections, and update $\mathcal{D}_c \!$\;
    }
    \For{$\tau_k$ collected before round $I$}{
        \For{$t\in\mathcal{N}_k$}{
            Query $\tilde{\mathbf{A}}_{k,t}^{I}
            \sim\pi^I(\cdot\mid\mathbf{O}_{k,t})$
            (Eq.~\eqref{eq:later_policy_query})\;
            Update the variance in Eq.~\eqref{eq:step_consistency} with
            $\tilde{\mathbf{A}}_{k,t}^{I}$\;
        }
    }
    Construct $\mathcal{D}_r$ by Eq.~\eqref{eq:relabeled_dataset}\;
    $\pi^{I+1}\leftarrow\pi^I$\;
    \For{$s=1,\ldots,N_{\text{training}}$}{
        Update $\pi^{I+1}$ by Eq.~\eqref{eq:full_objective} with $\mathcal{D}_c$
        and $\mathcal{D}_r$\;
    }
}
\end{algorithm}

\section{Experiment}

We design our experiments to answer three questions:
(1) How well does SeRA perform against the baselines in online and offline
learning (Section~\ref{sec:exp_comparison})?
(2) What contributes to the strong performance of SeRA
(Section~\ref{sec:exp_analysis})?
(3) Can SeRA be applied to a real-world task
(Section~\ref{sec:real_exp})?

\subsection{Experimental setup}
\label{sec:exp_setup}

\begin{figure}[t]
\centering
\includegraphics[width=\columnwidth]{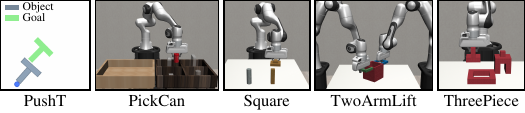}
\caption{Tasks for the simulation experiments.}
\label{fig:sim_tasks}
\end{figure}

\paragraph{Tasks}
We evaluate SeRA on five simulated manipulation tasks and one real-robot
task. The simulated tasks (Fig.~\ref{fig:sim_tasks}) include a PushT task~\cite{chi2023diffusion}, three
tasks from the robosuite benchmark~\cite{zhu2020robosuite} (PickCan, Square,
and TwoArmLift), and the ThreePiece assembly task from
MimicGen~\cite{mandlekar2023mimicgen}. All simulated tasks are evaluated in
both online and offline learning. The real-robot task is
described in Section~\ref{sec:real_exp}.

\paragraph{Simulated teacher}
For repeatable comparisons, we follow SDP~\cite{Li2026SDP} and use a simulated
teacher whose actions are optimal. The teacher monitors the robot and
intervenes once the distance between the robot action and the teacher action
exceeds a threshold. The teacher then controls the robot for $2T$ steps, where
$T$ is the chunk length, and returns control to the robot. As training
progresses, the teacher lets the robot act longer before monitoring again,
mimicking a human teacher who intervenes less often as the robot improves.

\paragraph{Online and offline learning}
In online learning, each method follows the IIL
loop (Section~\ref{sec:prelim_iil}) and collects its own data with the simulated
teacher. Each round consists of an interaction phase, in which the robot
collects one trajectory, followed by $N_{\text{training}}$ policy updates. To
reduce the computational cost, all methods start from the same SDP policy and
dataset, obtained by running this loop with SDP from scratch for 50 rounds.
Each method then collects data with its own policy and trains with its own
objective for another 250 rounds.
In offline learning, all methods train on the same dataset, collected by an
SDP agent during online learning. This setting separates how each method
learns from how its data was collected. Here, a round consists of
$N_{\text{training}}$ policy updates without data collection. We pre-train an SDP policy on this dataset for 150
rounds, and each method then continues from this checkpoint for
another 300 rounds.

\paragraph{Baselines}
All baselines are built on SDP and differ in how they use autonomous robot
experience:
(1) SDP~\cite{Li2026SDP} learns only from
complete teacher-labeled chunks;
(2) SDP+IWR~\cite{mandlekar2020iwr} imitates all autonomous robot experiences;
(3) SDP+CEILing~\cite{2022_RAL_Correct_me_If_Wrong} uses the final episode
reward as evaluative feedback and imitates robot actions from successful
trajectories;
(4) SDP+Sirius~\cite{liu2023sirius} uses intervention-guided weighting, which
assigns zero weight to pre-intervention actions;
(5) SDP+Pseudo-labeling applies the same cross-round relabeling and
consistency filtering as SeRA with its own checkpoints, but imitates
accepted relabels as exact targets ($r^N=0$).
IWR, CEILing, and Sirius likewise imitate the selected robot actions as exact
targets.
Except for SDP, all baselines also use the partial human actions, as does
SDP+StepMask, the variant of SeRA without relabeling (Section~\ref{sec:exp_analysis}).
We omit the prefix SDP+ when the context is clear. Hyperparameters are listed
in Appendix~\ref{sec:hyperparameters}.

\paragraph{Evaluation}
For each task and setting, we run three trials with different seeds. In each
trial, we save checkpoints every 5 of the 300 training rounds and evaluate
the last 21 checkpoints, each on 10 episodes with distinct initial states and
evaluation seeds. We report the success rate averaged over these checkpoints,
as the mean and standard deviation over seeds.

\providecommand{\sd}[1]{{\scriptsize$\pm$#1}}
\begin{table*}[t]
\centering
\caption{Success rate (\%) in simulation, averaged over the last 21 checkpoints
(mean $\pm$ std over seeds). Online, each method collects its own data with
the simulated teacher; offline, all methods train on the same dataset.}
\label{tab:sim_results}
\scriptsize
\setlength{\tabcolsep}{1.8pt}
\begin{tabular*}{\textwidth}{@{\extracolsep{\fill}}l*{10}{c}@{}}
\toprule
 & \multicolumn{5}{c}{Online} & \multicolumn{5}{c}{Offline} \\
\cmidrule(lr){2-6}\cmidrule(lr){7-11}
Method & PushT & PickCan & Square & TwoArmLift & ThreePiece
 & PushT & PickCan & Square & TwoArmLift & ThreePiece \\
\midrule
SDP & 57.9\sd{2.4} & 97.9\sd{1.7} & 91.7\sd{3.7} & 96.7\sd{2.7} & 94.0\sd{3.1} & 86.5\sd{0.7} & \textbf{99.8}\sd{0.3} & 90.8\sd{1.0} & 87.6\sd{0.8} & 93.3\sd{0.5} \\
SDP+IWR & 41.2\sd{6.4} & 84.0\sd{0.3} & 69.5\sd{0.0} & 90.0\sd{2.5} & 65.4\sd{5.1} & 52.1\sd{5.7} & 95.6\sd{1.2} & 76.0\sd{3.1} & 94.8\sd{1.7} & 80.2\sd{3.1} \\
SDP+Sirius & 34.8\sd{4.0} & 87.1\sd{5.4} & 77.4\sd{1.0} & 95.0\sd{1.7} & 62.1\sd{5.5} & 49.4\sd{3.6} & 95.9\sd{1.5} & 81.3\sd{2.9} & 95.6\sd{1.2} & 78.1\sd{3.7} \\
SDP+CEILing & 52.6\sd{5.7} & 95.2\sd{2.0} & 83.8\sd{4.7} & 97.9\sd{3.0} & 87.0\sd{2.1} & 72.4\sd{2.1} & 98.7\sd{1.1} & 85.4\sd{3.1} & 97.0\sd{1.0} & 93.2\sd{2.3} \\
SDP+Pseudo-labeling & 70.5\sd{5.4} & 73.1\sd{19.9} & 75.6\sd{13.1} & 87.4\sd{7.1} & 89.3\sd{3.0} & 66.4\sd{1.0} & 11.9\sd{10.9} & 65.0\sd{26.6} & 92.9\sd{0.0} & 37.6\sd{30.7} \\
\midrule
SDP+StepMask (ours) & 71.0\sd{3.9} & 98.6\sd{0.5} & 92.1\sd{1.0} & \textbf{99.5}\sd{0.0} & 96.3\sd{1.5} & 87.5\sd{2.6} & \textbf{99.8}\sd{0.3} & 93.3\sd{1.6} & 96.5\sd{0.9} & 97.1\sd{0.8} \\
SeRA (ours) & \textbf{79.8}\sd{2.8} & \textbf{99.8}\sd{0.3} & \textbf{96.2}\sd{0.5} & 99.4\sd{0.3} & \textbf{98.4}\sd{0.7} & \textbf{92.0}\sd{3.5} & \textbf{99.8}\sd{0.3} & \textbf{96.3}\sd{1.5} & \textbf{98.0}\sd{0.8} & \textbf{98.9}\sd{0.7} \\
\bottomrule
\end{tabular*}
\end{table*}

\subsection{Comparison with baselines}
\label{sec:exp_comparison}

Table~\ref{tab:sim_results} summarizes the results of online and offline learning
in simulation.

\subsubsection{SeRA consistently outperforms the baselines}
SeRA achieves the highest success rate among all baselines on every task,
in both online and offline learning. Compared with SDP, which uses no
autonomous robot experience, SeRA improves the success rate on every task except
offline PickCan, where both methods reach 99.8\%. The gain is largest on
online PushT and offline TwoArmLift, and small where SDP is near the
ceiling, as on PickCan. In contrast,
each baseline that uses autonomous robot experience falls below SDP on at
least four of the five tasks in each setting. SDP+Pseudo-labeling, which
imitates accepted relabels as exact targets, even falls far below SDP on most
tasks, especially in offline learning; we discuss this degradation in
Section~\ref{sec:exp_analysis}.
These results show that set-valued supervision of policy relabels is more
effective at leveraging autonomous robot experience than imitating historical
robot actions or exact relabels.

\subsubsection{Imitating historical robot actions can reduce task success}
IWR and Sirius fall below SDP on every online task and on every offline task
except TwoArmLift (Table~\ref{tab:sim_results}). CEILing performs best of the
three, but exceeds SDP only on TwoArmLift. Thus, retaining historical robot actions as imitation
targets can reduce task success, even when these actions are selected by
task success or intervention context. One possible explanation is that this
supervision reinforces suboptimal actions, which may occur without triggering
interventions or within successful trajectories. SeRA instead relabels
autonomous robot experience with later policies and uses the
historical robot actions only to scale the desired sets
(Section~\ref{sec:set_target}).

\begin{table}[t]
\centering
\caption{Component ablation of SeRA in offline learning. Success rate
(\%), mean $\pm$ standard deviation over three seeds, using the evaluation
protocol of Table~\ref{tab:sim_results}. ``w/o filtering'' removes checkpoint
consistency filtering; ``w/o set-valued targets'' sets the relabel radius ratio to
$r^N=10^{-4}$, yielding approximately point-valued supervision;
``w/o relabeling'' uses only retained human supervision (StepMask).
Results for StepMask and SeRA are repeated from
Table~\ref{tab:sim_results}.}
\label{tab:ablation}
\footnotesize
\setlength{\tabcolsep}{2pt}
\renewcommand{\arraystretch}{1.15}
\begin{tabular}{@{}lcccc@{}}
\toprule
Variant & PushT & Square & TwoArmLift & ThreePiece \\
\midrule
\rowcolor{gray!12}
SeRA & \textbf{92.0}\sd{3.5} & \textbf{96.3}\sd{1.5} & \textbf{98.0}\sd{0.8} & \textbf{98.9}\sd{0.7} \\
w/o filtering & 84.5\sd{2.4} & 31.4\sd{6.7} & 26.7\sd{0.0} & 9.5\sd{8.8} \\
w/o set-valued targets & 78.1\sd{9.5} & 59.4\sd{21.5} & 94.1\sd{2.2} & 36.3\sd{21.7} \\
w/o relabeling & 87.5\sd{2.6} & 93.3\sd{1.6} & 96.5\sd{0.9} & 97.1\sd{0.8} \\
\bottomrule
\end{tabular}
\end{table}

\subsection{Analysis of SeRA}
\label{sec:exp_analysis}

Table~\ref{tab:ablation} reports ablations of SeRA in offline
learning.

\paragraph{Relabeling complements partial human supervision}
Without relabeling, StepMask retains partial human actions through the step-masked
denoising loss and matches or exceeds SDP on every task in both settings
(Table~\ref{tab:sim_results}). Relabeling further improves success on all
four tasks in Table~\ref{tab:ablation}. Online, SeRA also exceeds StepMask
except on TwoArmLift, where both are near the ceiling. These comparisons
show that relabeling improves the policy beyond learning from partial human
actions alone.

\paragraph{Consistency filtering improves learning from relabels}
Removing consistency filtering lowers success from 96.3\% to 31.4\% on Square
and from 98.9\% to 9.5\% on ThreePiece (Table~\ref{tab:ablation}). The
unfiltered variant underperforms StepMask on all four tasks, showing that
unfiltered relabel supervision can impair learning even with set-valued
targets. One possible explanation is that filtering excludes unreliable
relabels that would otherwise impair training.

\paragraph{Set-valued relabel targets improve task success}
With filtering retained, reducing the relabel radius ratio from $r^N=1$ to
$10^{-4}$ yields approximately point-valued targets and lowers success on all
four tasks, most sharply on Square and ThreePiece (Table~\ref{tab:ablation}).
The exact-target baseline, SDP+Pseudo-labeling ($r^N=0$), also
underperforms SeRA on every task in both settings
(Table~\ref{tab:sim_results}). These comparisons support set-valued relabel
supervision in the evaluated settings. This pattern is consistent with confirmation
bias: relabel errors can enter later checkpoints and be reinforced by
subsequent relabeling. Relaxed matching may reduce this reinforcement.
Together, these ablations show that both consistency filtering and set-valued
targets contribute to SeRA's performance on the evaluated offline tasks.

\subsection{Real-robot experiments}
\label{sec:real_exp}

We evaluate SeRA and its variant without relabeling, StepMask, on a
real-world tomato-picking task. We compare them with SDP and three baselines
that imitate historical robot actions: IWR, CEILing, and Sirius.

\begin{figure}[t]
\centering
\includegraphics[width=0.48\textwidth]{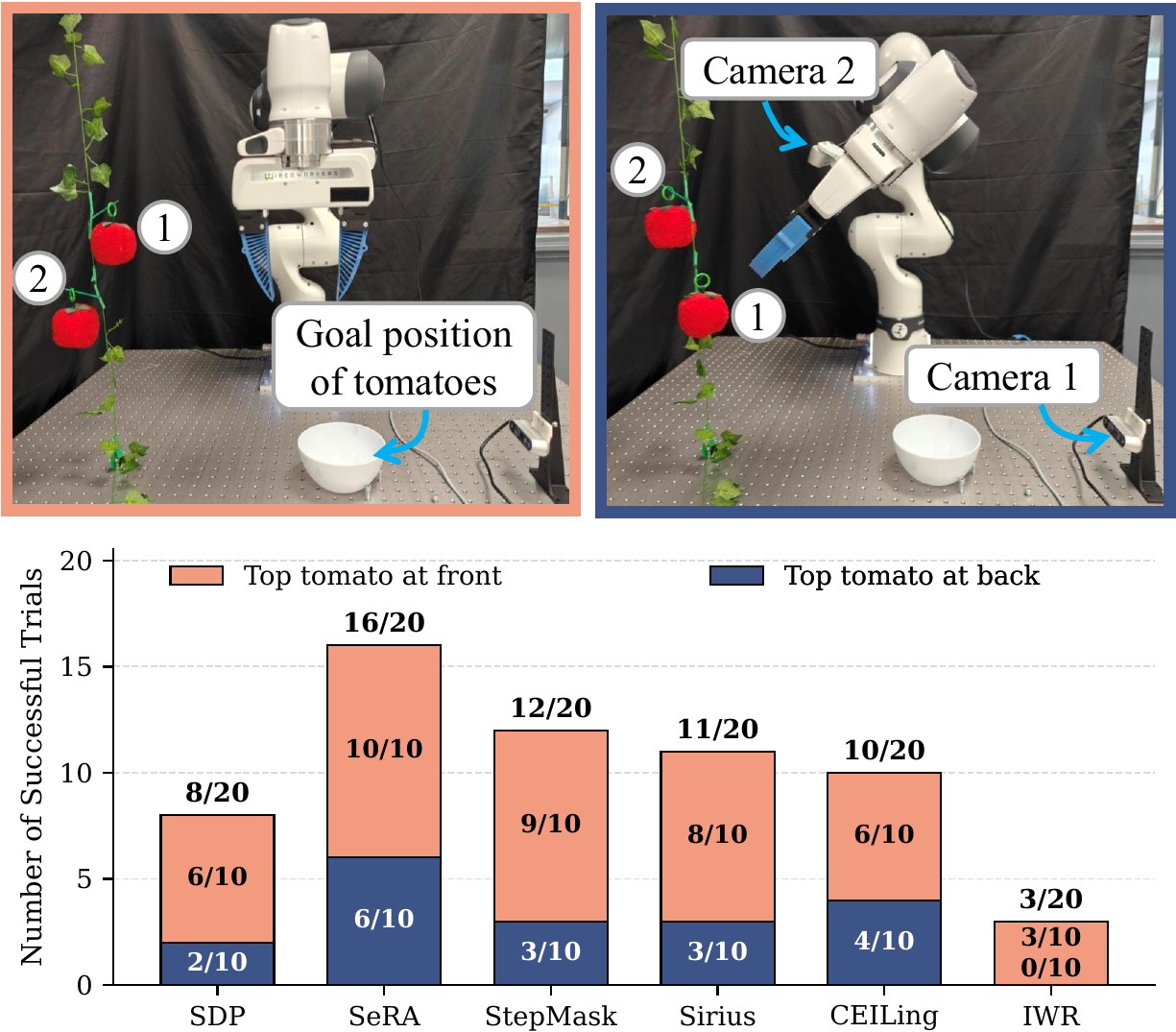}
\caption{Real-robot tomato-picking task. Top: the two evaluated
configurations, with the top tomato at the front (left) or at the back
(right). Bottom: number of successful trials for each method, out of 10
trials per configuration.}
\label{fig:real_exp_results}
\end{figure}

\paragraph{Task and evaluation}
The robot picks two tomatoes from a plant and places them in a bowl
(Fig.~\ref{fig:real_exp_results}, top). A trial is successful if the robot places
both tomatoes in the white bowl. We evaluate two configurations, with the top
tomato at the front or at the back of the plant, and run 10 trials per
configuration for each method. The robot's initial pose and the tomato goal
position are fixed across trials, with small variations in the tomatoes'
initial states within each configuration. We evaluate the final training
checkpoint of each method. The robot operates at a control frequency of
10~Hz. At each step, the observation includes RGB
images from the two cameras shown in Fig.~\ref{fig:real_exp_results} and the
robot's end-effector pose. Each single-step action consists of an
end-effector pose in the fixed world coordinate frame and a gripper command.
The human teacher provides demonstrations and
interventions with a SpaceMouse.

\paragraph{Training protocol}
We pre-train an SDP policy~\cite{Li2026SDP} on 12{,}000 demonstration samples
and deploy it on the robot, during which a human teacher provides corrective
interventions. The collected data contain 5{,}700 samples of human corrections and
19{,}200 samples of autonomous robot experience, including 4{,}591 chunks with partial
human actions. Each method is then trained offline, as in
Section~\ref{sec:exp_setup}, on the aggregated dataset of demonstrations,
corrections, and the autonomous robot experience that the method uses. To match the
computational budget, every method is trained for 14~hours on an NVIDIA A40
GPU.

\paragraph{Use of autonomous robot experience}
Table~\ref{tab:real_data} reports the samples of autonomous robot experience
used by each method at the end of training. All counts include the 4{,}591 chunks with partial
human actions. These chunks retain their human
supervision and use historical robot actions (IWR, CEILing, and Sirius) or
accepted relabels (SeRA) at unlabeled steps. SeRA combines the retained
human actions and accepted relabels in the step-masked denoising loss, whereas StepMask uses only partial human actions.

\paragraph{Results}
SeRA achieves the highest success rate in both configurations, with the
same human feedback and training budget as the other methods
(Fig.~\ref{fig:real_exp_results}, bottom). Across the 20 trials per method,
SeRA reaches 80\% success, compared with 40\% for SDP and 60\% for StepMask.
The gain over StepMask indicates that relabeling improves the policy beyond
retaining partial human actions. SeRA also outperforms Sirius, CEILing,
and IWR. IWR, which imitates all historical robot actions, performs worst
and fails in all trials with the top tomato at the back. Unlike on most
simulated tasks, Sirius and CEILing exceed SDP, but only by a few successful
trials. A possible explanation is that, in this task,
their selection rules discard some of the harmful robot actions that IWR
imitates, whereas the retained robot actions still limit their improvement
over SDP. These results show
that SeRA can learn from autonomous robot experience on a real robot
through relabeled, set-valued supervision rather than imitation of historical
robot actions.

\begin{table}[t]
\centering
\caption{Samples of autonomous robot experience used after real-robot training,
including chunks with partial human actions.}
\label{tab:real_data}
\footnotesize
\setlength{\tabcolsep}{2pt}
\begin{tabular*}{\columnwidth}{@{\extracolsep{\fill}}lrrrrr@{}}
\toprule
Method & IWR & CEILing & Sirius & StepMask & SeRA \\
\midrule
Samples & 19{,}200 & 16{,}800 & 15{,}200 & 4{,}591 & 12{,}905 \\
\bottomrule
\end{tabular*}
\end{table}

\section{Conclusion}
\label{sec:conclusion}

In this work, we introduced SeRA, a framework for learning from autonomous
robot experience alongside human corrections. SeRA uses later policy
checkpoints to relabel earlier observations and constructs set-valued action
targets around relabels selected by checkpoint agreement. A step-masked denoising loss
combines these targets with retained human corrections in partially supervised
action chunks. Experiments in online and offline simulation and on a real
robot show that SeRA improves policy performance over SDP and outperforms historical
robot-action imitation and exact relabel imitation in the evaluated settings.
Together, these findings show that set-valued relabeling extends learning
beyond human corrections to make effective use of autonomous robot experience.

\textit{Limitations and future work:}
SeRA assumes that the desired sets around accepted relabels contain optimal
actions. Checkpoint agreement cannot guarantee this assumption, since later
policies can share errors. The current method also uses a fixed radius ratio
for relabeled targets and discards relabels that fail the consistency test.
Future work could adapt desired sets using checkpoint variance and use
observations with high disagreement to guide additional queries or robot
resets near those states.

\ifanonymous\else
\section*{Acknowledgment}
This project is made possible by a contribution from the National Growth Fund program NXTGEN Hightech. We also acknowledge the Delft AI Cluster (DAIC) for providing computational resources. This work also used the Dutch national e-infrastructure with the support of the SURF Cooperative using grant no. EINF-17077.
\fi

\appendix


\subsection{Hyperparameters}
\label{sec:hyperparameters}

Table~\ref{tab:hyperparameters} lists the hyperparameters of SeRA. The same
values are used in online and offline learning. The remaining hyperparameters
follow SDP~\cite{Li2026SDP}.

\begin{table}[h]
\centering
\caption{Hyperparameters of SeRA. Values in parentheses apply to the
listed task or setting.}
\label{tab:hyperparameters}
\footnotesize
\begin{tabular}{@{}llc@{}}
\toprule
Symbol & Description & Value \\
\midrule
$T$ & Chunk length & 16 \\
$T_a$ & Executed steps per chunk & 8 \\
$\delta$ & Variance threshold & 0.01 (TwoArmLift: 0.001) \\
$M$ & Min.\ number of later checkpoints & 10 \\
$r^h$ & Radius ratio, human actions & 0.1 (real: 0.01) \\
$r^N$ & Radius ratio, accepted relabels & 1 \\
$N_{\text{training}}$ & Policy updates per round & 100 \\
\bottomrule
\end{tabular}
\end{table}

\paragraph{Weight $\beta$}
Instead of fixing $\beta$ in Eq.~\eqref{eq:full_objective}, we draw each
mini-batch from either $\mathcal{D}_c$ or $\mathcal{D}_r$ and select
$\mathcal{D}_r$ with probability
\[
    p_N\coloneqq
    \frac{|\mathcal{D}_r|}{|\mathcal{D}_c|+|\mathcal{D}_r|}.
\]

\bibliographystyle{IEEEtran}
\bibliography{references}

\end{document}